%% file: iwinac2022.tex
\documentclass[runningheads]{llncs}
\usepackage[T1]{fontenc}
\usepackage{draftwatermark}
\SetWatermarkText{draft}
\SetWatermarkColor[gray]{0.9}
\SetWatermarkScale{2}

\usepackage{graphicx}

\usepackage{xcolor}
\usepackage{epsfig} 
\usepackage{mathptmx} 
\usepackage{times} 
\usepackage{amsmath} 
\usepackage{amssymb}  
\usepackage{listings}
\usepackage{pythonhighlight}
\usepackage{hyperref}

\usepackage{tikz,xcolor,hyperref}

\definecolor{lime}{HTML}{A6CE39}
\DeclareRobustCommand{\orcidicon}{
	\begin{tikzpicture}
	\draw[lime, fill=lime] (0,0) 
	circle [radius=0.16] 
	node[white] {{\fontfamily{qag}\selectfont \tiny ID}};
	\draw[white, fill=white] (-0.0625,0.095) 
	circle [radius=0.007];
	\end{tikzpicture}
	\hspace{-2mm}
}
\foreach \x in {A, ..., Z}{\expandafter\xdef\csname orcid\x\endcsname{\noexpand\href{https://orcid.org/\csname orcidauthor\x\endcsname}
			{\noexpand\orcidicon}}
}

\input{PDDL}
\definecolor{backcolour}{rgb}{0.95,0.95,0.92}
\definecolor{codegreen}{rgb}{0,0.6,0}
\definecolor{stringcolor}{rgb}{240,0,0}
\lstdefinestyle{myStyle}{
    backgroundcolor=\color{backcolour},   
    commentstyle=\color{codegreen},
    stringstyle=\color{stringcolor},
    basicstyle=\ttfamily\footnotesize,
    breakatwhitespace=false,         
    breaklines=true,                 
    keepspaces=true,                 
    numbers=left,
    stepnumber=2,
    numbersep=5pt,                  
    showspaces=false,                
    showstringspaces=false,
    showtabs=false,                  
    tabsize=2,
}
\definecolor{darkgreen}{RGB}{0,120,0}
\definecolor{darkred}{RGB}{145,0,0}

\usepackage[ruled,vlined]{algorithm2e}

\newcommand*{\kant}{\texttt{KANT}}

\begin{document}
\title{KANT: A tool for Grounding and   Knowledge Management}

\def\mycmd{1} 

\if\mycmd1
{

%
%
\author{Miguel Á. González-Santamarta \inst{1}\orcidA{} \and
Francisco J. Rodríguez-Lera \inst{1}\orcidB{} \and
Francisco Martín \inst{2}\orcidC{} \and
Camino Fernández \inst{1}\orcidD{} \and 
Vicente Matellán \inst{1}\orcidE{}}
\authorrunning{M.A. González Santamarta et al.}
%
\institute{¹ Universidad de León, León , España \\
\email{\{mgons, fjrodl, cferl, vmato\}@ unileon.es} \\
²Universidad Rey Juan Carlos, Madrid, España\\
\email{\{francisco.rico\}@urjc.es }}
}
\fi
\maketitle              
\begin{abstract}
The intelligent robotics community usually organizes knowledge into symbolic and sub-symbolic levels. These two levels establish the set of symbols and rules for manipulating knowledge based on their (symbol system  - dictionary). Thus, the correspondences - Grounding or knowledge representation- require specific software techniques for anchoring continuous and discrete state variables between these two levels. This paper presents the design and evaluation of an Open Source tool called \kant\space (Knowledge mAnagemeNT) to let different components of the system architecture controlling the robot query, save, edit, and delete the data from the Knowledge Base without having to worry about the type and the implementation of the source data. Using \kant, components managing subsymbolic information can smoothly interact with symbolic components. Besides, implementation mechanisms used in \kant, such as the use of in-memory and non-SQL databases, improve the performance of the knowledge management systems in ROS middleware, as shown by the evaluations presented in this work.
\end{abstract}

\section{Introduction}

Wide-spreading of service robotics applications in the real world means that robots have to perform long-term tasks in highly dynamic environments. Planning this long-term tasks, in particular when interaction with humans is involved, requires acquiring, and updating symbolic representations. Implementing this process is a complex task that involving different techniques from the Artificial Intelligence field, and using Software Engineering technologies adapted to robotics.

Software development for robots is nowadays built using specific middlewares, ROS (Robotic Operating System~\cite{Quigley:2020}) being the {\em de facto} standard. Using these tools, modules in charge of the symbol generation are usually built as components that can be reused among different projects. For instance, one component can be built to generate grounded information about the presence of a specific set of objects (cups, glasses, etc.) in the scene from images, another one can assess the presence of people from LIDAR readings, etc. These components need to share the symbolic information that they generate into the symbol-based decision making system of the robot, that are usually based on planning techniques. 

Symbolic planning systems are "abstract, explicit deliberation process that chooses and organizes actions" \cite{ghallab2004} for changing the state of the robot or the environment. In order to do that, they use symbolic representation of the capabilities of the robot given by the designers (or learnt), and the symbolic representation of their knowledge about the environment obtained by the grounding components previously mentioned. The {\em de facto} standard way of representing all this knowledge is using high-level language as PDDL~\cite{pddl} (Planning Domain Definition Language).

The robot PDDL domain and problem definitions need to be maintained and updated over time. There is not a standard way to manage and store PDDL knowledge for robotic middleware, in particular for ROS. As a result, there are several different methods such as single or multiple nodes for maintaining Domain and PDDL, thus, it is complicated to compare or integrate PDDL mechanisms.  

%




\kant\space intends to solve this problems by simplifying the technical process of interacting with knowledge storage. In particular, facilitating the use of high-level languages such as PDDL in "ROS 2" middleware, easing its use in real robots. 
It lets developers recognize PDDL objects by their names and attributes, providing an abstraction to manage robot knowledge in practically any scenario, independently of the storage technologies. 

The inherent problems of PDDL manipulation can be solved using software design patterns.
This paper presents \kant\footnote{https://github.com/uleroboticsgroup/kant}, as an approach for the encapsulation of the PDDL language through four software design patterns in order to simplify the access to PDDL information. Particularly, this work proposes the use of Data Transfer Object (DTO), Data Access Object (DAO), Abstract Factory and Factory Method design patterns. As a result, scalability, flexibility and reusability are improved. Besides, developers do not have to worry about the implementation of the storage implementation of the Knowledge Base. 

%

The rest of this paper is organized as follows. Section \ref{sec:relatedwork} explores the state of the art. Section \ref{sec:KM} describes the engine from a high level and development level. Section~\ref{sec:Experiment} report the description and results of our experiment.  The paper shows the discussion of our approach in Section~\ref{sec:discussion} and conclusions are presented in Section \ref{sec:conclusions}.

\section{Related Work} 
\label{sec:relatedwork}

There are several software alternatives facing the problem of using different software components for storing the knowledge of robots using PDDL. ROSPlan~\cite{kn:rosplan} is the most popular planning solution in the ROS ecosystem. It provides a ROS node-based "Knowledge Base" that allows managing knowledge expressed as PDDL. ROSPlan uses the ROS communication interfaces, which involves creating the proper messages and using the right clients for each case. For instance, querying and updating knowledge have different messages, and PDDL types and PDDL propositions has different ROS services. These services manage the data of the ROS node. This can be considered as an in-memory solution which is inadequate for long-term accountability processes. Our approach proposes a disconnected solution that allows its integration not only with ROS but also with databases for storing knowledge. 
Besides, ROSPlan is available just for ROS 1, and there are no plans to migrate it to ROS 2. Thus, \kant\space has been naively developed for ROS 2.

PlanSys2 \cite{PlanSys2} is the alternative to ROSPlan in ROS 2. It offers advanced approaches, such as the use of behavior trees to execute the generated plans, as well as command-line options to interact with the knowledge base, which is composed of two ROS nodes, one for the PDDL domain and another one for the PDDL problem. One of the limitations of PlanSys2 is that PDDL support is not complete. It supports PDDL 2.1. Also, for now, it only supports two plan solvers POPF and TFD. Besides, it also maintains a certain grade of coupling to ROS due to the use of its communications mechanisms to manipulate the PDDL knowledge.

The work presented in \cite{7829420} uses the Entity-Component-System (ECS) pattern to create actions similar to the PDDL actions. Then, those actions can be translated into PDDL to be used by a planner. 
Nevertheless, a complete example of the use of PDDL is not presented and it is only mentioned that PDDL actions can be obtained.

The authors in \cite{10.1145/3276954.3276961} introduce "Tool Ontology and Optimization Language" (TOOL), which is based on OWL 2 \cite{motik2009owl}. 
TOOL has a domain-specific language (DSL) whose elements are offered as Kotlin types that are created using the Factory Method design pattern. The elements of this DSL are translated into an OWL ontology, which is classified. Finally, the ontology is used to generate the PDDL domain that can be solved. This means that an extra step for knowledge manipulation is required.


Other alternatives such as OARA architecture~\cite{oara} defines a specific language (DSL) to create robot skills that are the actions a robot needs to perform to achieve a mission. Then, using this language, the PDDL domain can be generated to be used in a planner. However, the OARA approach based on DSL requires knowledge of this language which makes it highly coupled to the architecture itself and \kant\space would be used standalone just using in Python.



\section{Knowledge Management Engine}
\label{sec:KM}

\kant\space presents a straightforward mecanism for managing the Knowledge Base, an engine supported by software patterns for encapsulating knowledge, particularly expressed in PDDL, and allowing access and manipulation from different robot software components. This approach enhances reusability, scalability and inter-operability of robot components managing knowledge (from grounding to processing components) in long-term tasks. 

\begin{figure}[t]
 \centering
 \includegraphics[width=0.45\columnwidth]{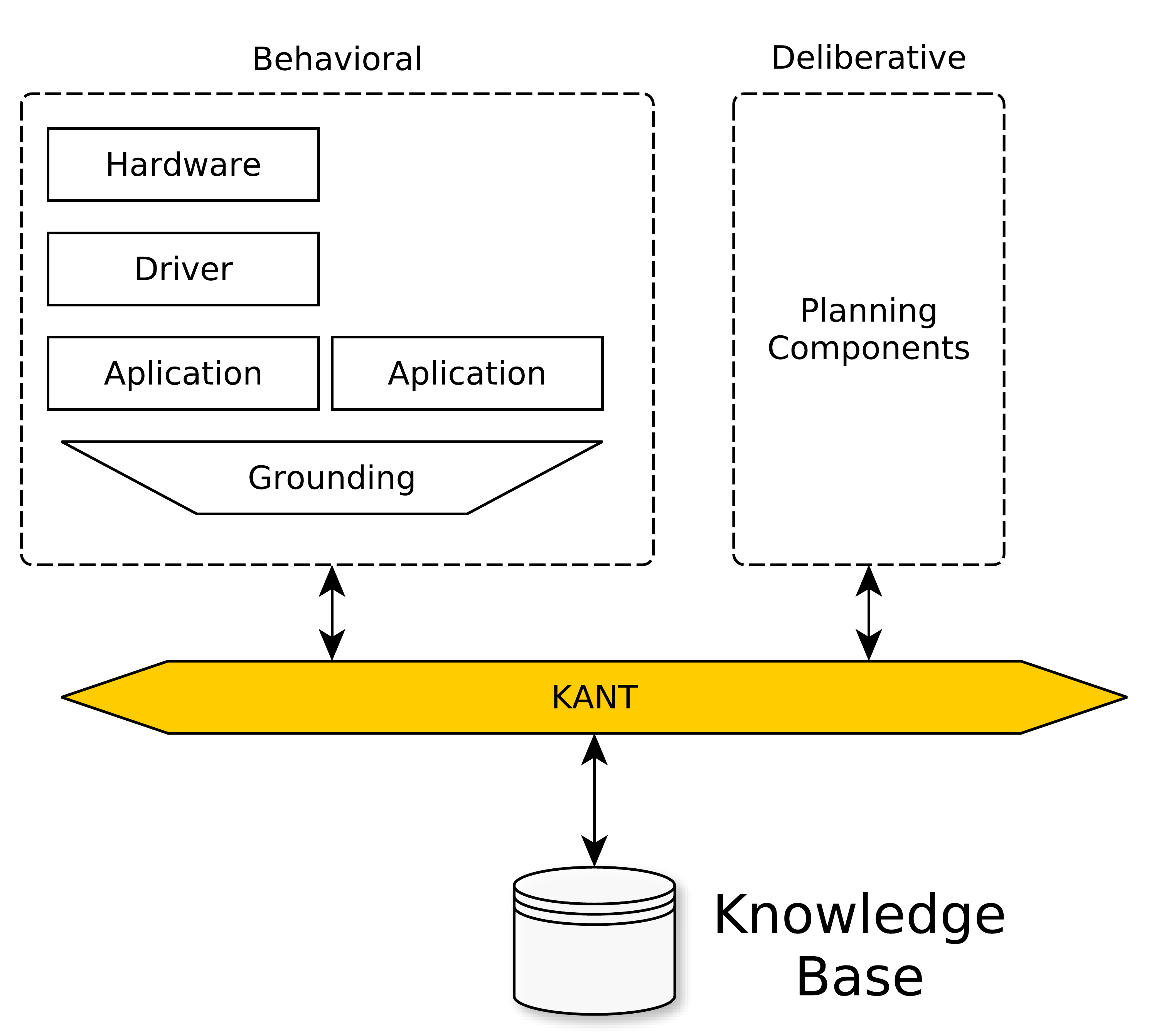}
 \caption{Kant in a generic robotic architecture.}
 \label{fig:kant_arch_1}
\end{figure}

\kant\space is represented as a yellow intermediate in figure~\ref{fig:kant_arch_1}. Any robot component can use it to access the Knowledge Base. As a result, any component can query and modify knowledge in the same way and without caring about how it is stored. Besides, updating the knowledge of a robot from sensors data is more simple thanks to it.

Representing knowledge using software patterns should be encouraged by new software developers focused on robotics. DTO, which stands for "Data Transfer Object"~\cite{Monday:2003}, aims to transfer the data between software components. DTO pattern is used to encapsulate PDDL information that can be shared between robotic components.

DTO pattern is used in \kant\space to create the classes that encapsulate the PDDL information that afterwards is shared between robotic components. DTO is based on creating private attributes that represent the data. These attributes must be accessed and edited with the getter, setter and constructor functions. Following this, the 


DAO is an abbreviation for Data Access Object\cite{Nock:2004}. It aims to use a software component to abstract and encapsulate all access to the data source. The DAO handles the connection to the data source to obtain and store data so it should encapsulate the logic for retrieving, saving and updating data in your data storage (a database, a file system, whatever). In this case, the data source will be the Knowledge Base where PDDL is stored. There can be a DAO family for each storage type treated.



%
%


\subsection{PDDL Abstractions in \kant}
\label{dto:overallOverview}

The PDDL planning domain model used in this research is the PDDL 2.1\cite{FoxPDDL21:2003} and is defined by the next tuple $\langle T, O, P, P, A \rangle$:
\begin{itemize}
    \item Types: represent something and restrict what can form the parameters of an action.
    \item Objects: is a set of typed objects, allow to declare objects that are present in the world with a twofold: name-type.
    \item Predicates: represent something that can be true in the world. It can involve several objects.
    \item Propositions: is a set of features that describe current the world.
    \item Actions: define the transformation between the states of the world.
    
\end{itemize}

The PDDL elements have been encapsulated into DTO elements. Following this, we have developed a DTO for each PDDL element. The structure of each PDDL element modeled with the DTO pattern is the following:
\begin{itemize}
    \item Types: A type is composed of only one string attribute, which represents its name. 
    \item Objects: An object is composed of one Type DTO attribute, which represents its PDDL type; and one string attribute, which represents its name. 
    \item Predicates: A predicate is composed of one string attribute, which represents its name; and one Type DTO List attribute, which represents its arguments. 
    \item Propositions: A proposition is composed of one string attribute, which represents its name; and one Object DTO List attribute, which represents its PDDL objects. It also has a boolean attribute, which represents if it is a goal.
    \item Actions: An action is composed of the following attributes:
    \begin{itemize}
        \item A string attribute, which represents its name.
        \item A boolean attribute, which represents if it is a durative action.
        \item An integer attribute, which represents its duration.
        \item An Object DTO List attribute, which represents its parameters.
        \item A Condition/Effect DTO List attribute, which represents its conditions. The Condition/Effect DTO is similar to the Proposition DTO but it also has a string attribute, which represents the moment the condition or effect must occur; and a boolean attribute, which represents if it is a negative condition or effect.
        \item A Condition/Effect DTO List attribute, which represents its effects.
    \end{itemize}
\end{itemize}


On the other hand, the DAO instances can be used to interact with the PDDL knowledge in the Knowledge Base. 
Three main functions can be used:
\begin{itemize}
    \item \emph{get}: this function returns a DTO from the Knowledge Base. It provides access to the current PDDL that the robot has to all robot components. In the case of PDDL types, objects, predicates and actions, this function takes their names to search for them. However, in the case of the PDDL propositions, there are three get functions: get\_by\_predicate, to search for a list of propositions with a given predicate name; get\_goals, to search for the propositions that are goals; and get\_no\_goals, to search for the propositions that are not goals.
    \item \emph{save}: this function is used to save a DTO. If the knowledge exists, it is updated. As a result, all robot components can produce knowledge for the robot.
    \item \emph{delete}: this function is also implemented in all DAO and is used to delete a DTO from the Knowledge Base.
\end{itemize}

Since the roboticist would have many DAO families such as sql-database, non-sql databases, in memory solutions for storing robot knowledge, \kant\space deploys the Abstract Factory Pattern. With this pattern, it is encapsulated different factories that have something in common, that is the creation of DAO. 

\begin{figure}[t]
 \centering
 \includegraphics[width=0.7\columnwidth]{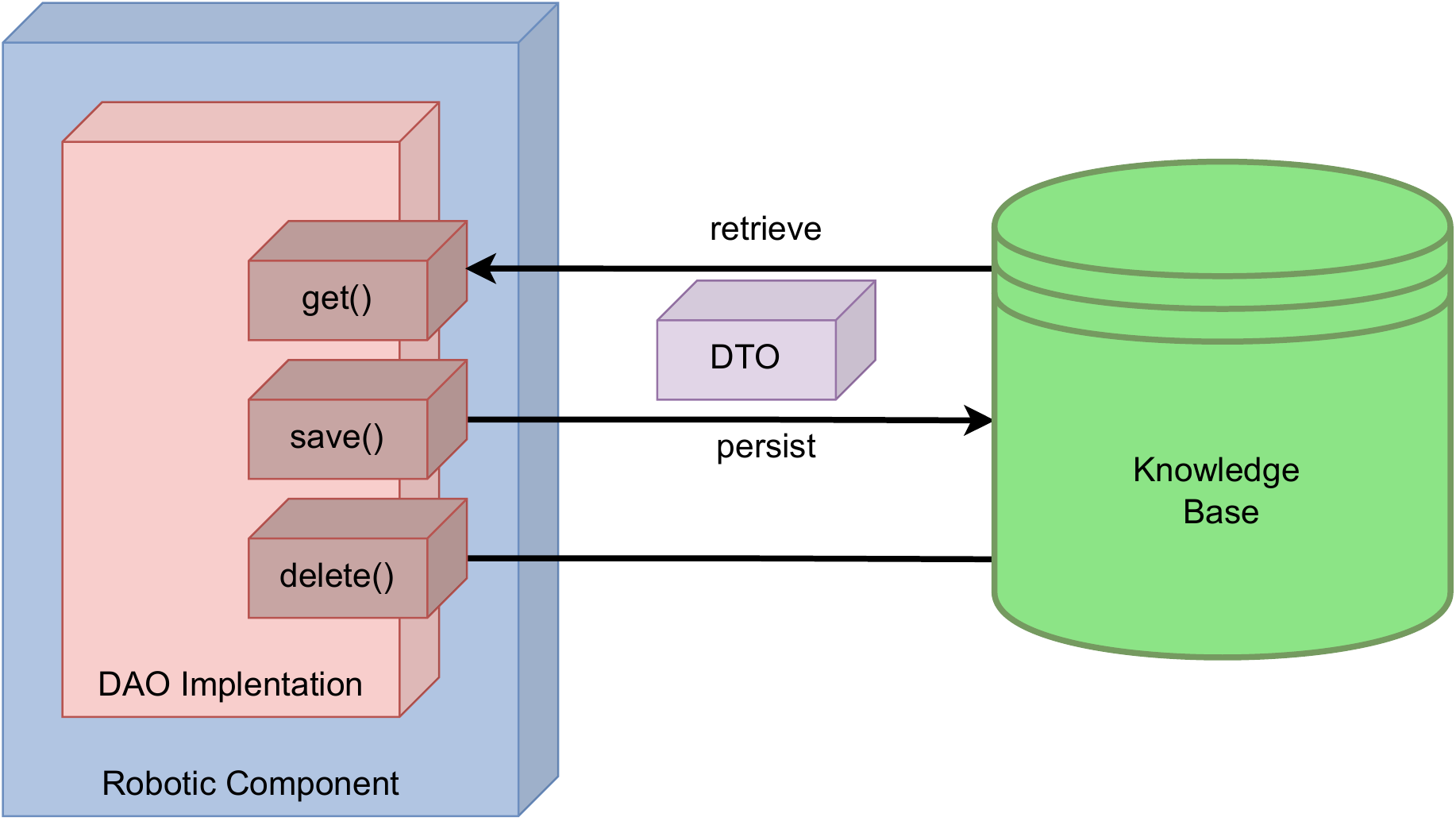}
 \caption{KANT Architecture}
 \label{fig:dao_arch}
\end{figure}

When combining DTO and DAO patterns, a robot component can query, save, edit and delete the data from the Knowledge Base without having to worry about the type and the implementation of the data source. That means that the DAO implementation acts like an interface between the source code of the developers and the Knowledge Base. Besides, the DTO pattern allows encapsulating the PDDL independently of the type of storage used. 
The resulting architecture is presented in Figure~\ref{fig:dao_arch}. As a result, developers can manipulate and manage knowledge from source code enhancing reusability, scalability and flexibility.


\input{KantInAction}
 \vspace{-0,2in}
\section{Evaluation Procedure}
\label{sec:Experiment}



For evaluating our proposal, it is proposed a high level task extracted and adapted from ERL-SciRoc 2019 Deliver coffee shop orders chapter~\cite{Basiri2018}. 
This task presents a robot that will assist people in a coffee shop. The robots will take care of customers, by taking orders and bringing objects to and from  tables.

 Because this competition domain is quite deterministic and dynamic, these tasks are 
 representative of a real-world performance. During performance, the robot needs a considerable number of objects, actions, and propositions that are queried, created, edited and deleted during execution.

These tasks are discretized and it is propsed temporal metric for modeling the enviornment. This experimental process evaluates the completion time of these five consecutive tasks that includes PDDL manipulation using KANT.

\begin{enumerate}
    \item Time to reset PDDL (TtR): this is the time spent to delete all the knowledge from the Knowledge Base. 

    \item Time to load initial PDDL (TtL): this is the time spent to load the initial PDDL elements. 
    
    \item Time to check tables (TtC): this is the time spent to simulate the task of checking several tables. 

    \item Time to serve an Order (TtS): this is the time spent to simulate the task of serving an order. 
    
    \item Time to guide a Person (TtG): this is the time spent to simulate the task of guiding a person. 

\end{enumerate}

It is implemented two DAO families to be able to store PDDL: 1) MongoDB \cite{bradshaw2019MongoDB} and 2) Knowledge Base in-memory using ROS 2. These families are deployed together with the three-layer cognitive architecture MERLIN~\cite{Gonzalez:2020} and everything has been simulated using a ROS 2 Foxy environment. An extended explanation of the experiment is presented in the public \href{https://github.com/uleroboticsgroup/kant}{GitHub Repository}.

\subsection{Results} 
After running 3000 times each test in a regular computer (laptop with 32 GB RAM and an Intel(R) Core(TM) i7-8750H CPU @ 2.20GHz), the results presented in Table~\ref{table:MERLIN} and Table~\ref{table:MONGO} are obtained. 

DAO in-memory approach presents an execution time of almost one hundred twenty-two minutes. The mean time of each iteration is 2.432 seconds with a median of 2.385 seconds. The time to load the PDDL gets 52 minutes, resetting knowledge takes 13 minutes, and the other tasks are performed in 23, 18 and 14 minutes in TtC, TtS and  TtG tasks. 

\begin{table}[!htb]
    \begin{minipage}{.55\linewidth}
     \caption{Manipulating PDDL using in-memory.}
\label{table:MERLIN}
\resizebox{.95\textwidth}{!}{
\begin{tabular}{|l|l|l|l|l|l|l|}
\hline
 & TtR & TtL & TtC  & TtS & TtG & Total\\ \hline

Mean	&	0.263&	1.051&	0.465&	0.364&	0.29&	2.432	\\ \hline
Std Dev.	&	0.124&	0.285&	0.18&	0.152&	0.137&	0.446	\\ \hline

Min	&	0.039&	0.527&	0.218&	0.16&	0.114&	1.312	\\ \hline
Max	&	0.805&	3.041&	2.065&	1.618&	1.489&	4.722	\\ \hline
Sum	&	787.571&	3152.132&	1393.824&	1093.036&	869.431&	7295.995\\ \hline

\end{tabular}
}
    \end{minipage}%
    \begin{minipage}{.55\linewidth}
      \centering
       \caption{Manipulating PDDL using MongoDB.}
\label{table:MONGO}
\resizebox{.95\textwidth}{!}{
\begin{tabular}{|l|l|l|l|l|l|l|}
\hline
 & TtR & TtL & TtC  & TtS & TtG & Total\\ \hline

Mean	&	0.069&	0.612&	0.329&	0.288&	0.216&	1.514\\ \hline	
Std dev	&	0.002&	0.014&	0.008&	0.007&	0.006&	0.027	 \\ \hline

Min	&	0.063&	0.569&	0.299&	0.262&	0.194&	1.446	 \\ \hline
Max&		0.1	& 0.756&	0.411&	0.344&	0.264&	1.815	 \\ \hline
Sum&		206.676&	1835.754&	987.505&	865.309& 646.987&	4542.231 \\ \hline
\end{tabular}
}
    \end{minipage} 
\end{table}




DAO MongoDB family defines an execution time of seventy-six minutes for the same 3000 iterations. From the total time, 3.5 minutes were used for resetting the database. Approximately 31 minutes were devoted to loading the PDDL information. When dealing with the tasks we found that during the experiment, it spent 16 minutes for the Checking the tables task, then 14 minutes for the Serving an Order and finally 10 minutes for the time to guide a person. Tables  ~\ref{table:MERLIN} and ~\ref{table:MONGO} present the descriptive statistics of the overall experiment measured in seconds.  




\section{Discussion}
\label{sec:discussion}

The processes involved in knowledge management should present exactly what is happening behind the scenes in order to guarantee explainable 
%
%
 reasoning in robotics.
Besides, the use of ROS 2 as the based middleware, simplify the use of \kant\space in state of the art robots and amplifies its impact on the overall community. This would avoid to break the rule "Don't reinvent the Wheel".

The results collected also present the quality difference when the model is managed by the right approach. 
\kant\space is a tool for easily using different storage systems. The figures present the differences between storages, where the same experiment would be carried out in a smaller period of time.	
%
%
For instance TtR spends a 70\% fewer time in MongoDb, consuming 35\% lower of time for the full experiment.  However, the inclusion of databases are not always possible, leaving the selection at the discretion of the researcher deploying the solution that fits better to their robotic scenario.  
 
\section{Conclusions}
\label{sec:conclusions}

In this paper, a novel approach for manipulating PDDL using software design patterns has been proposed. This tool simplifies the ways of interacting with PDDL knowledge in run-time from a software engineer perspective. The evaluation has shown that \kant is an efficient tool usuful for easing the management of PDDL by robotic developers in a standardized way. Authors believe that the models based on well known software patterns together with the implemented solution in Python 3 would simplify the use of PDDL as the common representation for symbolic planning in robotics.

As a future work, we plan to evaluate the performance of this approach in robotics competition scenarios. We are also interested on analyzing from the educational point of view if new robotic researchers can interact with PDDL more comfortably using \kant. Finally, the current implementation needs to be upgraded with more PDDL elements such as Numeric Fluents, in order to encompass modern versions of PDDL. 

\subsubsection{Acknowledgements} The research described in this work has been partially funded by the Kingdom of Spain under grant XXXX.

%
%
%
\bibliographystyle{splncs04}
\bibliography{references}

\end{document}

%% file: PDDL.tex
\lstdefinelanguage{PDDL}
{
  sensitive=false,    
  morecomment=[l]{;}, 
  alsoletter={:,-},   
  morekeywords={
    define,domain,problem,not,and,or,when,forall,exists,either,
    :domain,:requirements,:types,:objects,:constants,
    :predicates,:action,:parameters,:precondition,:effect,
    :fluents,:primary-effect,:side-effect,:init,:goal,
    :strips,:adl,:equality,:typing,:conditional-effects,
    :negative-preconditions,:disjunctive-preconditions,
    :existential-preconditions,:universal-preconditions,:quantified-preconditions,
    :functions,assign,increase,decrease,scale-up,scale-down,
    :metric,minimize,maximize,
    :durative-actions,:duration-inequalities,:continuous-effects,
    :durative-action,:duration,:condition
  }
}

%% file: KantInAction.tex
\subsection{Implementation Overview}
\label{sec:example}

This section presents a practical guide for applying KANT to manage PDDL knowledge. To this end, the software approach for managing a simple robot task, the navigation between waypoints, using Code Listings (CL onwards), is schemed. 

\subsubsection{Task and Knowledge}

A simple robot task is proposed as an example to illustrate how the developer interacts with PDDL. The task consists of moving a robot called rb1; from its current way-point, named wp1; to another way-point, labeled as wp2 in a period of time of 10 seconds. 


\subsubsection{Translating the task to PDDL from the source code}

The roboticist and the AI researcher need to focus on a single point, generating and managing a PDDL Domain file.
The PDDL domain file presents two types, robot and wp (waypoint); one predicate, robot\_at that implies if a robot is in a waypoint (wp); and one durative-action, navigation, that models the navigation between waypoints for a time period and involves a robot in a new waypoint from a previous waypoint. Besides, there are three objects in the initial problem that are a robot named rb1 and two wp, wp1 and wp2. Finally, the goal that wants to be achieved is to have the rb1 robot in wp2.

\subsubsection{Using PDDL from the source code}

Firstly, the DTO classes are used. These classes store the PDDL information. As a result, the PDDL types previously mentioned are instantiated in Python using the \textit{PddlTypeDto} class with the name of each type. This is presented in CL \ref{lst:python_types_dto}. 

\vspace{-0,2in}
\input{Python/types}

\vspace{-0,2in}
Predicates are created with the \textit{PddlPredicateDto} class. The robot\_at predicate is presented in \ref{lst:python_predicates_dto}. This predicate is created with its name and the \textit{PddlTypeDto} instances of the necessary PDDL types.

\input{Python/predicates}

\vspace{-0,2in}
The creation of a PDDL object is similar to predicate creation. As it is presented in \ref{lst:python_objects_dto}, a PDDL object can be created using the \textit{PddlObjectDto} class that needs its name and a \textit{PddlTypeDto} instance. 

\vspace{-0,2in}
\input{Python/objects}

\vspace{-0,2in}
The PDDL proposition is created using the \textit{PddlPropositionDto} class with a \textit{PddlPredicateDto} instance and its \textit{PddlObjectDto} instances. Besides, a proposition can be marked as a goal using the is\_goal argument (CL \ref{lst:python_propositions_dto}).
\vspace{-0,2in}
\input{Python/propositions}

\vspace{-0,2in}
PDDL actions  are created using the \textit{PddlActionDto} class. As it is shown in CL \ref{lst:python_navigation_action_dto}, parameters, conditions and effects must be created. 

\input{Python/actions}

The action parameters are \textit{PddlObjectDto} while conditions and effects are created with a new class, \textit{PddlConditionEffectDto}. The \textit{PddlConditionEffectDto} instances are created like \textit{PddlPropositionDto} instances but with the possibility of selecting the time, if the action is durative. Conditions and effects can also be negative. The available times for conditions and effects are AT\_START, AT\_END and OVER\_ALL. Then, an action can be created with its name, its parameters, a list of \textit{PddlObjectDto} instances; and its conditions and effects, two lists of \textit{PddlConditionEffectDto} instances. In addition, an action can be durative or not using the durative argument, which is True by default.\\

%% file: Python/types.tex
\begin{lstlisting}[
    float=!htb,
    caption={Python PDDL Types DTO Example.},
    label={lst:python_types_dto},
    basicstyle=\scriptsize,
    numberfirstline=true,
    language=Python]
robot_type  = PddlTypeDto("robot")
wp_type     = PddlTypeDto("wp")
\end{lstlisting}

%% file: Python/predicates.tex
\begin{lstlisting}[
  float=!htb,
  caption={Python PDDL Predicates DTO Example.},
  label={lst:python_predicates_dto},
  basicstyle=\scriptsize,
  language=Python]
robot_at = PddlPredicateDto("robot_at", [robot_type, wp_type])
    
\end{lstlisting}

%% file: Python/objects.tex
\begin{lstlisting}[
  float=!htb,
  caption={Python PDDL Objects DTO Example.},
  label={lst:python_objects_dto},
  basicstyle=\scriptsize,
  language=Python]
rb1 = PddlObjectDto(robot_type, "rb1")
wp1 = PddlObjectDto(wp_type, "wp1")
wp2 = PddlObjectDto(wp_type, "wp2")

\end{lstlisting}

%% file: Python/propositions.tex
\begin{lstlisting}[
  float=!htb,
  caption={Python PDDL Propositions DTO Example.},
  label={lst:python_propositions_dto},
  basicstyle=\scriptsize,
  language=Python]
pddl_proposition_dto = PddlPropositionDto(robot_at, [rb1, wp1])

pddl_goal_dto = PddlPropositionDto(robot_at, [rb1, wp2], is_goal=True)

\end{lstlisting}

%% file: Python/actions.tex
\begin{lstlisting}[
  float=!htb,
  caption={Python PDDL Navigation Action DTO Example.},
  label={lst:python_navigation_action_dto},
  basicstyle=\scriptsize,
  language=Python]
r = PddlObjectDto(robot_type, "r")
s = PddlObjectDto(wp_type, "s")
d = PddlObjectDto(wp_type, "d")
    
condition_1 = PddlConditionEffectDto(robot_at, [r, s],
    time=PddlConditionEffectDto.AT_START)

effect_1 = PddlConditionEffectDto(robot_at, [r, s],
    time=PddlConditionEffectDto.AT_START,
    is_negative=True)

effect_2 = PddlConditionEffectDto(robot_at, [r, d],
    time=PddlConditionEffectDto.AT_END)

pddl_action_dto = PddlActionDto("navigation", [r, s, d],
    [condition_1], [effect_1, effect_2])

\end{lstlisting}

%% file: iwinac2022.bbl
\begin{thebibliography}{10}
\providecommand{\url}[1]{\texttt{#1}}
\providecommand{\urlprefix}{URL }
\providecommand{\doi}[1]{https://doi.org/#1}

\bibitem{Basiri2018}
Basiri, M., Piazza, E., Matteucci, M., Lima, P.: Rulebook of the european
  robotic league for consumer service robots (2018),
  \url{https://eu-robotics.net/robotics_league/erl-consumer}

\bibitem{bradshaw2019MongoDB}
Bradshaw, S., Chodorow, K., Brazil, E.: MongoDB: the Definitive Guide: Powerful
  and Scalable Data Storage. The expert's voice in open source, O'Reilly Media,
  Incorporated (2019), \url{https://books.google.es/books?id=ohGAvgAACAAJ}

\bibitem{kn:rosplan}
Cashmore, M., Fox, M., Long, D., Magazzeni, D., Ridder, B., Carrera, A.,
  Palomeras, N., Hurtós, N., Carreras, M.: {ROSplan}: Planning in the robot
  operating system. In: Proceedings International Conference on Automated
  Planning and Scheduling, ICAPS. vol.~2015, pp. 333--341 (01 2015)

\bibitem{oara}
Charles~Lesire, D.D., Grand, C.: Formalization of robot skills with descriptive
  and operational models. {IEEE} (2020)

\bibitem{7829420}
{Fischbach}, M., {Wiebusch}, D., {Latoschik}, M.E.: Semantic entity-component
  state management techniques to enhance software quality for multimodal
  vr-systems. IEEE Transactions on Visualization and Computer Graphics
  \textbf{23}(4),  1342--1351 (2017). \doi{10.1109/TVCG.2017.2657098}

\bibitem{FoxPDDL21:2003}
Fox, M., Long, D.: Pddl2. 1: An extension to pddl for expressing temporal
  planning domains. Journal of artificial intelligence research  \textbf{20},
  61--124 (2003)

\bibitem{10.1145/3276954.3276961}
Gavran, I., Mailahn, O., M\"{u}ller, R., Peifer, R., Zufferey, D.: Tool:
  Accessible automated reasoning for human robot collaboration. In: Proceedings
  of the 2018 ACM SIGPLAN International Symposium on New Ideas, New Paradigms,
  and Reflections on Programming and Software. p. 44–56. Onward! 2018,
  Association for Computing Machinery, New York, NY, USA (2018).
  \doi{10.1145/3276954.3276961}

\bibitem{ghallab2004}
Ghallab, M., Nau, D., Traverso, P.: Automated Planning: theory and practice.
  Elsevier (2004)

\bibitem{Gonzalez:2020}
Gonz{\'{a}}lez-Santamarta, M.{\'{A}}., Rodr{\'{\i}}guez-Lera, F.J.,
  {\'{A}}lvarez-Aparicio, C., Guerrero-Higueras, {\'{A}}.M.,
  Fern{\'{a}}ndez-Llamas, C.: {MERLIN} a cognitive architecture for service
  robots. Applied Sciences  \textbf{10}(17), ~5989 (aug 2020).
  \doi{10.3390/app10175989}

\bibitem{PlanSys2}
Mart{\'{\i}}n, F., Gin{\'{e}}s, J., Rodr{\'{i}}guez, F.J., Matell{\'{a}}n, V.:
  Plansys2: A planning system framework for ros2. In: {{IEEE/RSJ} International
  Conference on Intelligent Robots and Systems, {IROS} 2021, Prague, Czech
  Republic, September 27 - October 1, 2021}. {IEEE} (2021)

\bibitem{pddl}
McDermott, D., Ghallab, M., Howe, A., Knoblock, C., Ram, A., Veloso, M., Weld,
  D., Wilkins, D.: Pddl-the planning domain definition language  (1998)

\bibitem{Monday:2003}
Monday, P.B.: Implementing the data transfer object pattern. In: Web Services
  Patterns: Java™ Platform Edition, pp. 279--295. Springer (2003)

\bibitem{motik2009owl}
Motik, B., Patel-Schneider, P.F., Parsia, B., Bock, C., Fokoue, A., Haase, P.,
  Hoekstra, R., Horrocks, I., Ruttenberg, A., Sattler, U., et~al.: Owl 2 web
  ontology language: Structural specification and functional-style syntax. W3C
  recommendation  \textbf{27}(65), ~159 (2009)

\bibitem{Nock:2004}
Nock, C.: Data access patterns: database interactions in object-oriented
  applications. Addison-Wesley Boston (2004)

\bibitem{Quigley:2020}
Quigley, M., Conley, K., Gerkey, B., Faust, J., Foote, T., Leibs, J., Wheeler,
  R., Ng, A.Y., et~al.: Ros: an open-source robot operating system. In: ICRA
  workshop on open source software. vol.~3, p.~5. Kobe, Japan (2009)

\end{thebibliography}
